\documentclass[conference]{utils/IEEEtran}

\usepackage{cite} 
\usepackage[pdftex]{graphicx} 
\usepackage{amsmath} 
\usepackage{array} 
\usepackage[caption=false,font=footnotesize]{subfig} 
\usepackage{url} 
\usepackage{tikz}
\usepackage{float}
\usepackage{algorithm}
\usepackage{algpseudocode}

\graphicspath{{inc/graphics/}}

\begin{document}

\title{
  Deep Learning for Inferring the Surface Solar Irradiance from Sky Imagery
}

\author{
  \IEEEauthorblockN{
    Mehdi Zakroum\IEEEauthorrefmark{1},
    Mounir Ghogho\IEEEauthorrefmark{1}\IEEEauthorrefmark{3},
    Mustapha Faqir\IEEEauthorrefmark{1}
    and Mohamed Aymane Ahajjam\IEEEauthorrefmark{1}
  }
  \IEEEauthorblockA{
    \IEEEauthorrefmark{1}International University of Rabat, TICLab, Morocco
  }
  \IEEEauthorblockA{
    \IEEEauthorrefmark{3}University of Leeds, School of IEEE, UK
  }
  \IEEEauthorblockA{
    \texttt{\{mehdi.zakroum,mounir.ghogho,mustapha.faqir,aymane.ahajjam\}@uir.ac.ma}
  }
}

\maketitle


\begin{abstract}
  We present a novel approach to perform ground-based estimation and prediction
  of the surface solar irradiance with the view to predicting photovoltaic
  energy production. We propose the use of mini-batch \(k\)-means clustering to
  extract features, referred to as per cluster number of pixels (PCNP), from
  sky images taken by a low-cost fish eye camera. These features are first used
  to classify the sky as clear or cloudy using a single hidden layer neural
  network; the classification accuracy achieves 99.7\%. If the sky is
  classified as cloudy, we propose to use a deep neural network having as input
  features the PCNP to predict intra-hour variability of the solar irradiance.
  Toward this objective, in this paper, we focus on estimating the deep neural
  network model relating the PCNP features and the solar irradiance, which is
  an important step before performing the prediction task. The proposed deep
  learning-based estimation approach is shown to have an accuracy of 95\%.
\end{abstract}

\vspace{5mm}


\section{Introduction}
With the proliferation of photovoltaic (PV) energy systems, it is important to
develop a reliable energy management system for operators to optimize the
integration of PV energy into the electrical grid. Predicting the energy
generated by the PV installation is a key feature in such a system, since it
allows to detect faulty performance, to perform load scheduling, and in general
to make better operational decisions \cite{cococcioni2011}. Forecasting
surface solar irradiance (SSI) is the basis of forecasting the PV energy
because of the direct relation between these two.

If the sky is clear, physics-based prediction models perform well. Since some
of the parameters in these models may be difficult to obtain, data-driven
methods have been proposed to predict the SSI. For example, the Global
Horizontal Irradiance (GHI) was shown to be accurately predicted with the
nonlinear auto-regressive with exogenous inputs (NARX) model, having as input
past GHI values \cite{elhendouzi2016}.

Predicting the SSI in the case of a cloudy sky is much more challenging than in
the case of clear sky conditions. This is due to the non-stationarity of the
clouds introduced by the stochastic behaviour of the wind in the spatial (e.g.
height of the clouds) and temporal dimensions. To address this issue, two
categories of methods have been proposed in the literature: ground-based
methods and geostationary satellite-based methods. Satellite-based prediction
is primarily carried out using numerical weather prediction and satellite cloud
monitoring. However, this approach only provides rough estimation due to the
spatial and temporal resolution limitations of satellite images
\cite{Chow20112881}. Ground-based methods, using sky imaging, have the
potential to provide better performance, particularly for the prediction of
intra-hour variability of the irradiance \cite{chu2014}. This is the approach
adopted in this paper.  

Toward the objective of forecasting the SSI, in this paper, we propose to use
deep learning to accurately estimate the mapping function between the sky image
and the corresponding irradiance, which is an important step before addressing
the forecasting task.  The data set used in this paper consists of sky images,
taken with a low-cost fish eye camera, and their corresponding Global
Horizontal Irradiance (GHI). First, the size of the images is reduced by
clustering the colors of the images' pixels into a relatively small number of
color clusters; this is carried out using the mini-batch \(k\)-means clustering
algorithm.  The resulting segmented sky image is used to classify the sky as
either clear or cloudy; the classifier was learned using manually labeled
images and a single hidden layer neural network. Using the cloudy sky image
subset, a deep neural network is trained to model the relationship between the
sky images and the corresponding irradiances.

\section{Data set}
\begin{figure*}[!t]
  \centering
  \subfloat[Clear sky]{
    \includegraphics[width=1.7in]{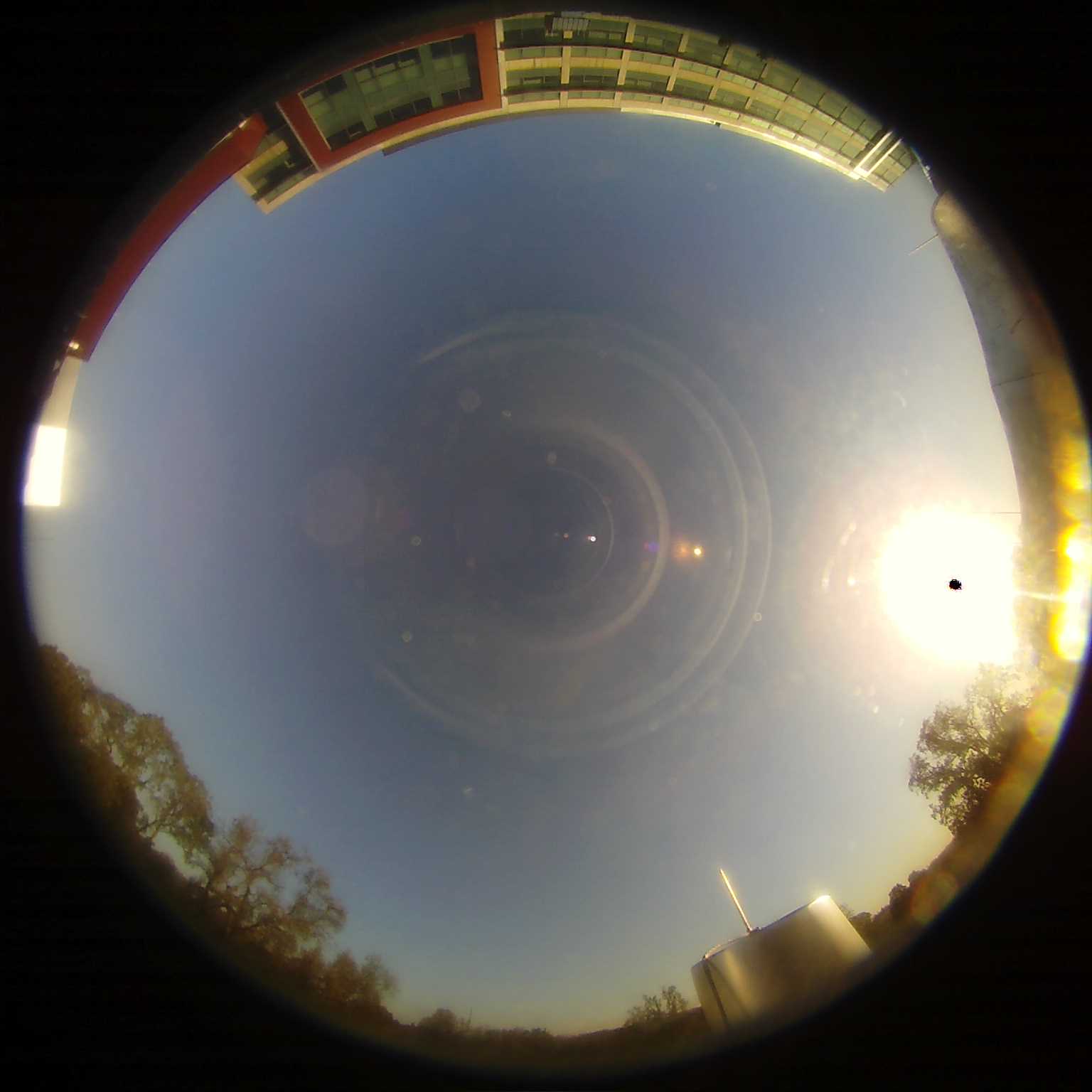}
    \label{fig:clear_sky}
  }
  \hfill
  \subfloat[Cloudy sky]{
    \includegraphics[width=1.7in]{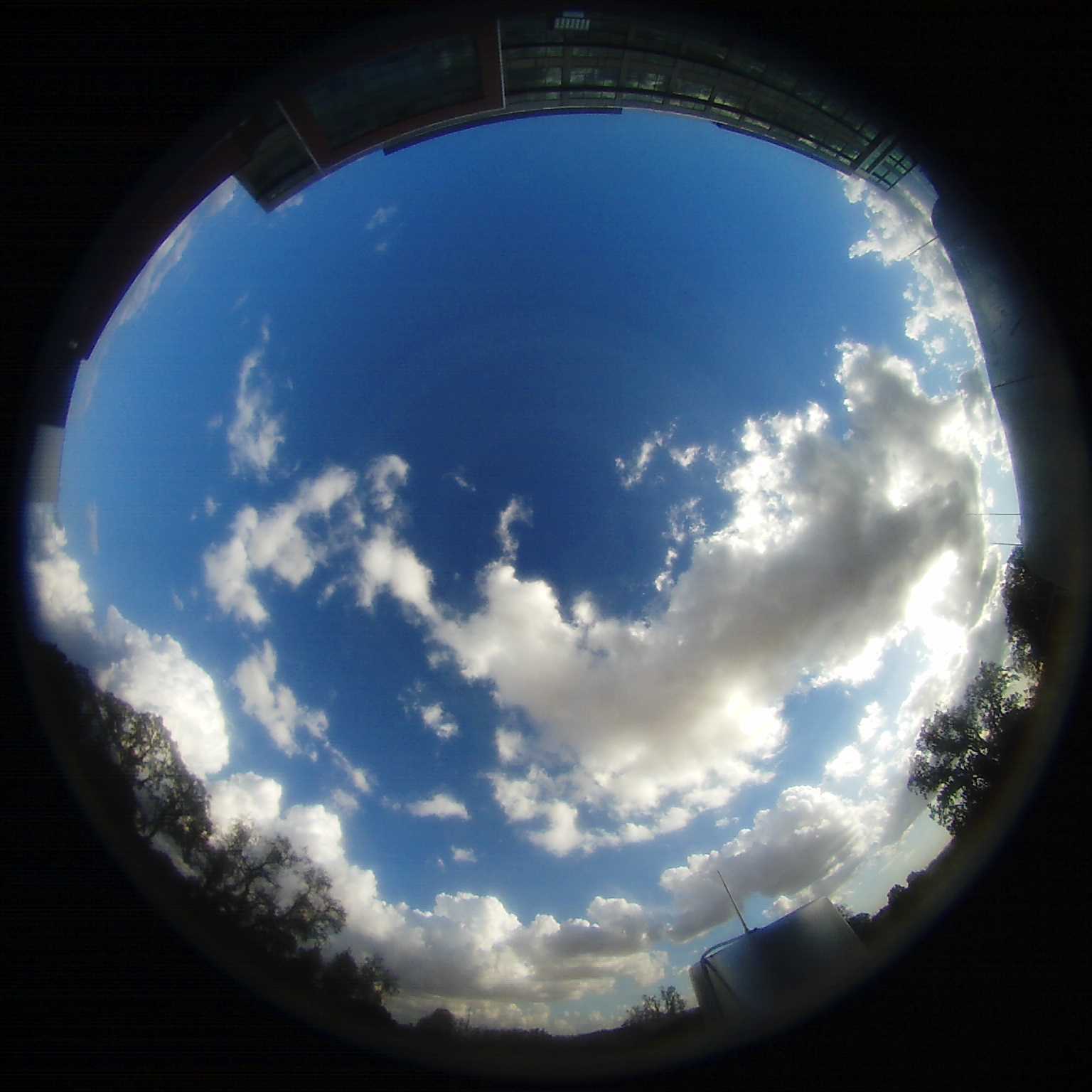}
    \label{fig:cloudy_sky}
  }
  \hfill
  \subfloat[Overcast sky]{
    \includegraphics[width=1.7in]{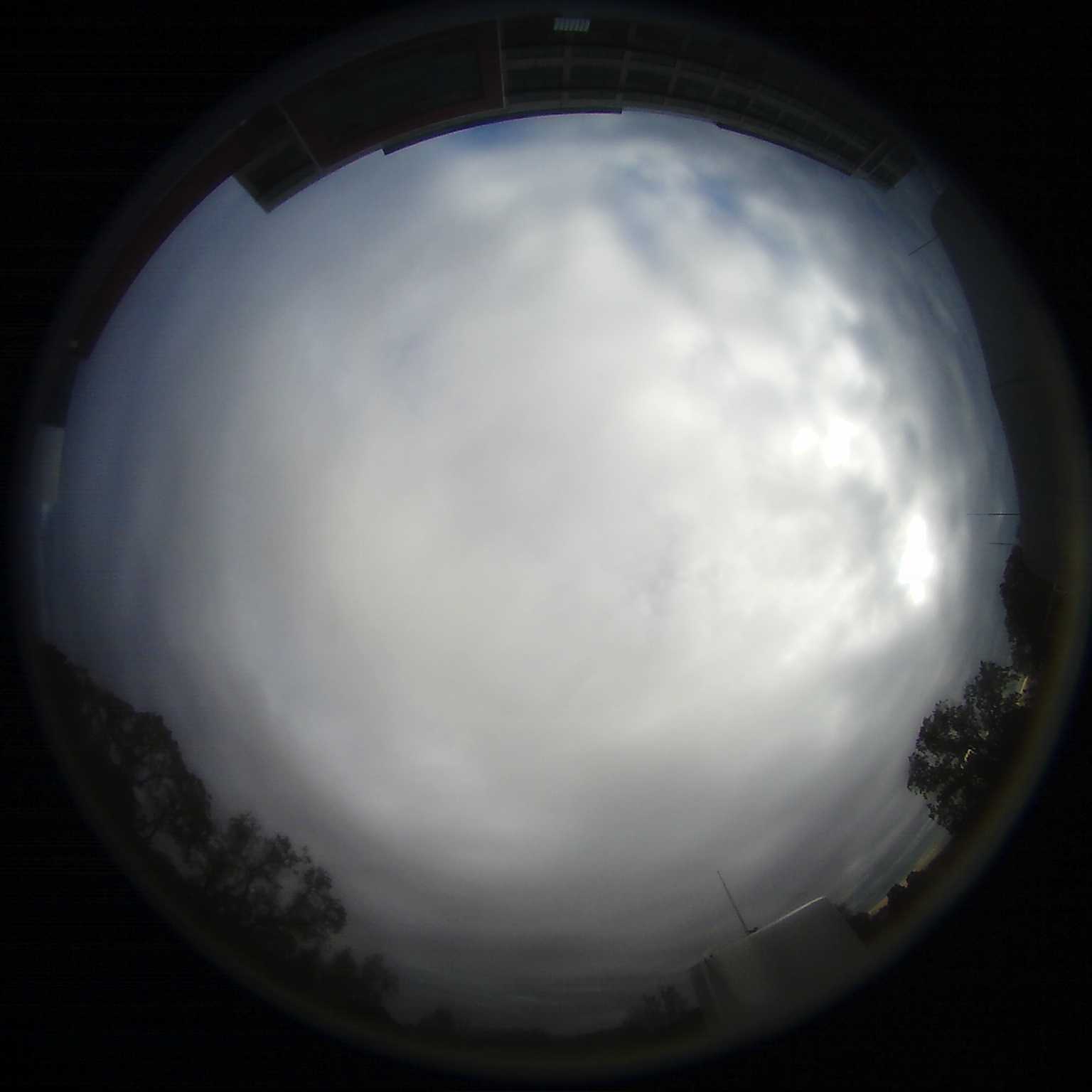}
    \label{fig:overcast_sky}
  }
  \caption{A sample of sky images with different weather conditions}
  \label{fig:folsom}
\end{figure*}

We use a data set consisting of RGB hemispheric sky images of size
\(512\times512\) pixels, taken with a Vivotek FE8174 fish-eye camera, and their
corresponding GHI, recorded with Kipp \& Zonen CMP11 pyranometer. The data set
was recorded during a period of three months (October, November and December of
2016) in Folsom, CA, USA. A record is taken every minute. More details about
this data set can be found in \cite{chu2014}. Figure~\ref{fig:folsom} shows a
sample of three sky images corresponding to different weather conditions.  

\section{Sky image segmentation and feature extraction}
Segmenting an image consists of reducing it to salient pixels that reflect its
global perceptual aspects. These representative pixels, also called
super-pixels, define the segments of a reduced image. Each segment has a label,
usually a color, and is present in one or more regions in the image. This task
provides a convenient representation of the image for further analysis. It also
serves to lower the computational cost because of the reduced number of values
taken by the channels of the color model in use.

Our goal is to simplify the representation of the RGB sky images without losing
much information about the three sky components, which are the clear sky, the
clouds and the solar disk. These three sky components are represented by
different sets of a relatively small number of RGB colors, with each one of
them contributing to the GHI with a certain weight. For instance, clouds might
have different levels of brightness, depending on the light they receive from
the sun, going from white near the solar disk to dark gray in the surroundings.
The intuition behind extracting different segments of the sky component is that
we could infer the GHI by knowing the number of pixels of each segments of the
segmented sky image. Indeed, a decrease in the GHI due the presence of clouds,
particularly when (partially) obscuring the sun, is manifested in the segmented
sky image by a reduction of the number of pixels present in the sun segments,
an increase of the number of pixels in the cloud segments, and a reduction of
the number of pixels representing the clear-sky.  

Clustering is one of the commonly used methods to segment an image. The most
widely used clustering algorithm is \(k\)-means. The statement of the problem is as
follows: given \(n\) pixels in the \(3\)-dimensional space defined by the R, G
and B channels (the axes of the space), and an integer \(k\), find a set of
\(k\) super-pixels, so as to minimize the mean squared distance between each
pixel and its nearest super-pixel. The caught super-pixels should be
representative of all the colors present in the sky images of the whole data
set, knowing that the colors vary with the time of day and the weather
conditions, as depicted in Figure~\ref{fig:folsom}.

\begin{figure*}[!t]
  \centering
  \subfloat[Clear sky]{
    \includegraphics[width=1.7in]{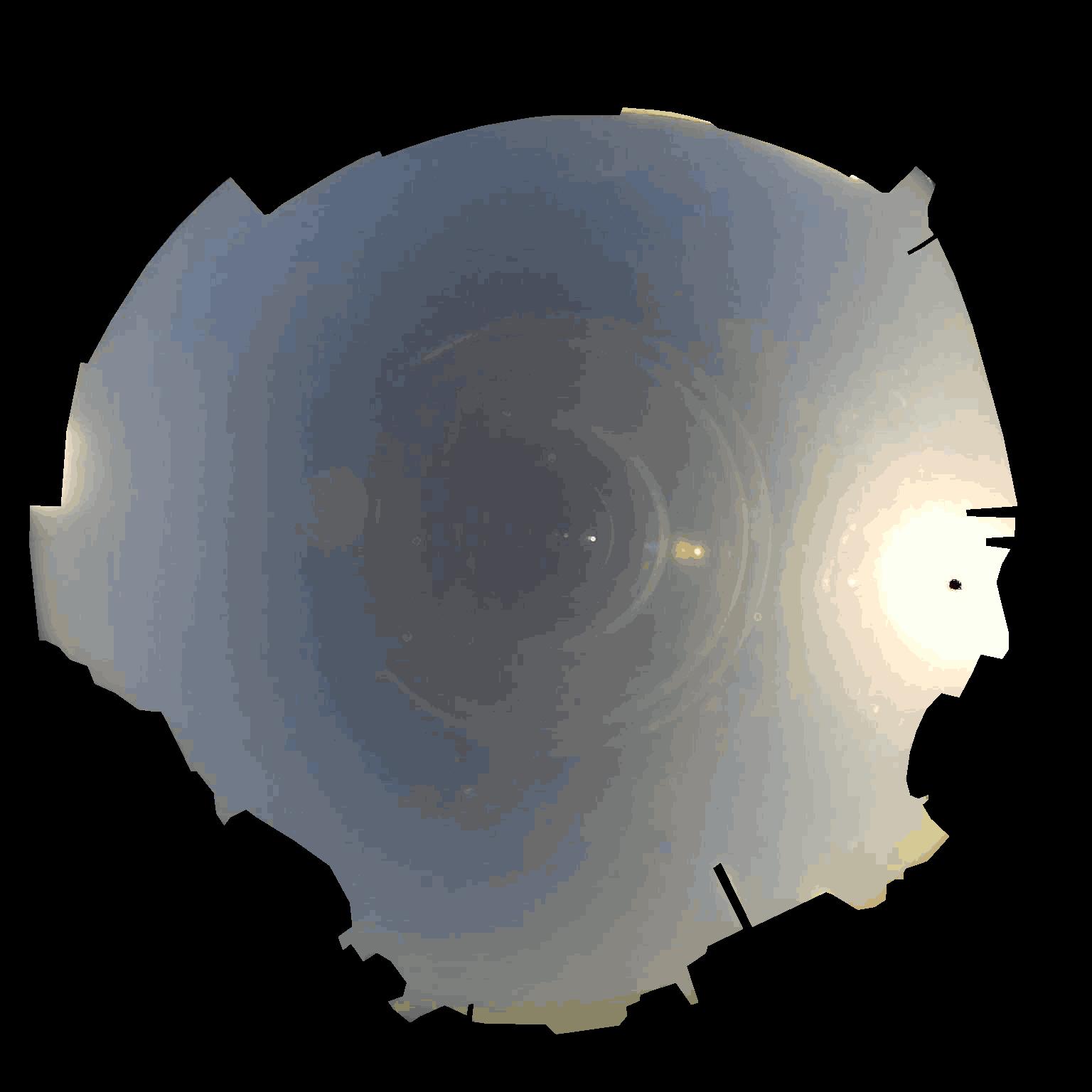}
    \label{fig:clear_sky_seg}
  }
  \hfill
  \subfloat[Cloudy sky]{
    \includegraphics[width=1.7in]{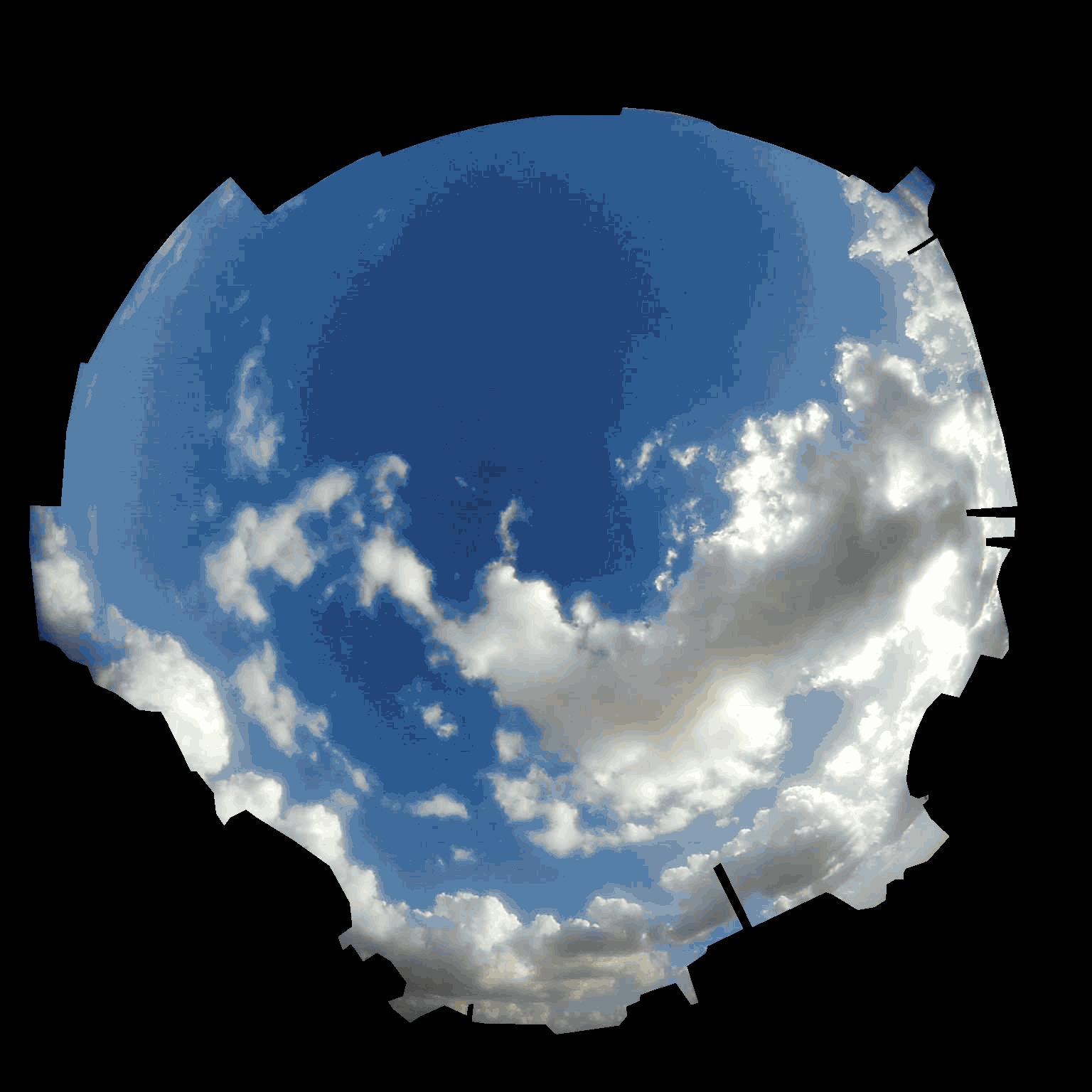}
    \label{fig:cloudy_sky_seg}
  }
  \hfill
  \subfloat[Overcast sky]{
    \includegraphics[width=1.7in]{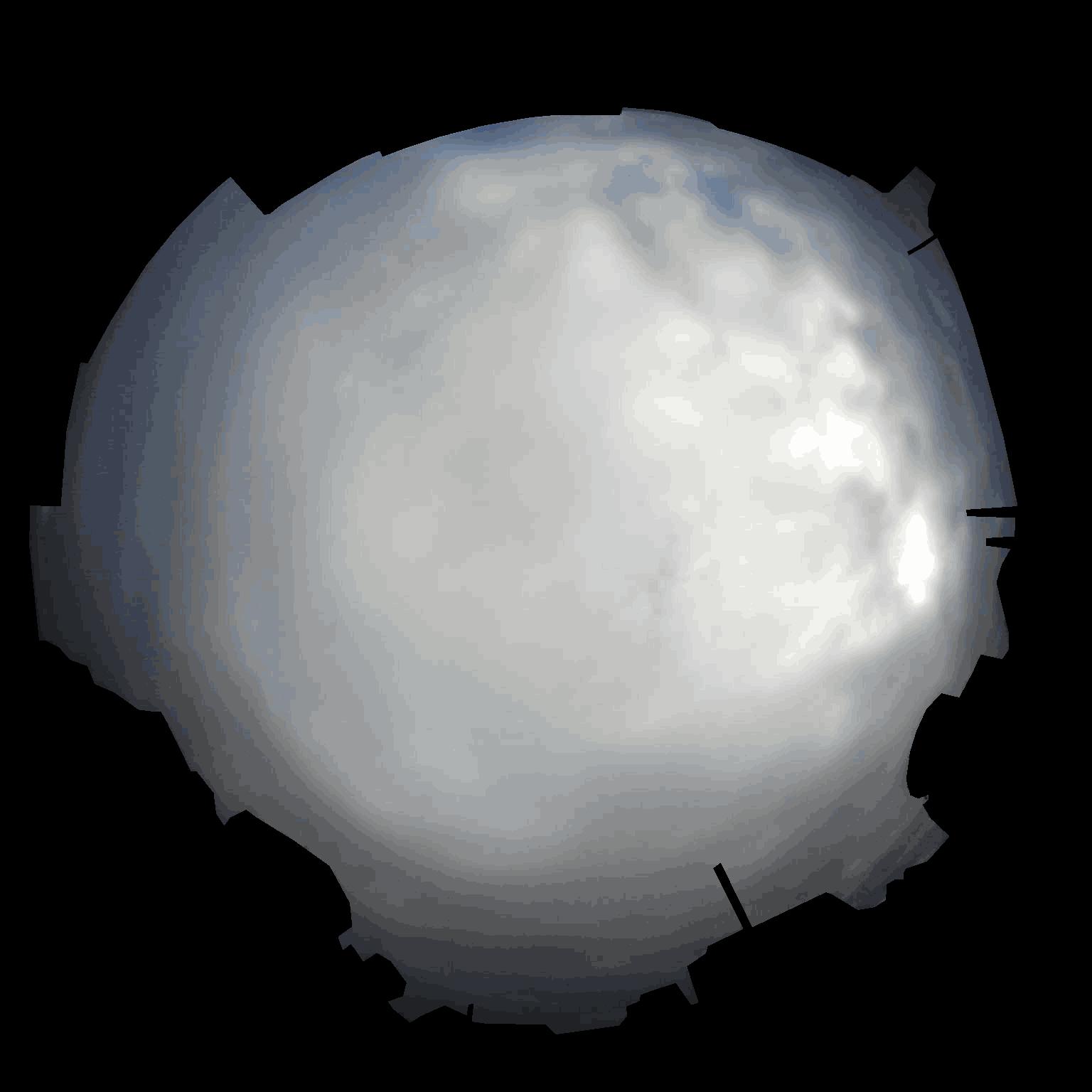}
    \label{fig:overcast_sky_seg}
  }
  \caption{Segmented versions of the sky images shown in Figure~\ref{fig:folsom}
  using a trained \(k\)-means model (64 clusters). The images belong to the
  testing set and were never visited during the training phase. The black
  pixels depict the applied mask.}
  \label{fig:folsom_seg}
\end{figure*}

The standard version of the \(k\)-means algorithm, first proposed by
Lloyd~\cite{lloyd1982}, is suitable for relatively small data sets (which is
not our case), because it requires to load all the images in memory before
processing. A workaround to this memory limitation problem is to train the
\(k\)-means algorithm on a manually selected set of sky images that catch most of
the typical colors representing the sky components. A better solution is to use
the stochastic gradient descent implementation of the \(k\)-means algorithm. To
reduce the stochastic noise and thus allow convergence to better centroids, a
mini-batch version of the stochastic gradient descent \(k\)-means algorithms was
proposed in \cite{sculley2010}. Figure ~\ref{fig:folsom_seg} shows clustered
sky images using a trained \(k\)-means model with \(k = 64\). It is worth pointing
out that before running the \(k\)-means algorithm, a mask hiding the surrounding
components which are not part of the sky is applied to the images in order to
intercept exclusively sky pixels.

After training the mini-batch \(k\)-means algorithm on the entire set of sky
images, we perform the segmentation using the resultant centroids. Then, for
each segmented image, we extract the number of pixels present in each segment.
We refer to these features in the following sections as \textit{per cluster
number of pixels} (PCNP).

\section{Sky image classification}
Many studies in the literature used NARX neural networks to forecast the PV
energy \cite{cococcioni2011,elhendouzi2016}. Even though this model showed
excellent ability in predicting the solar irradiance (and directly the PV
energy) under clear sky conditions, there are no studies, to our knowledge,
that cover the problem of the noise generated by the clouds that affects the
solar irradiance, making the NARX model performance to degrade. The aim of this
study is to enhance the prediction capabilities of such a model by taking into
account the configuration of the sky. In order to decide about the prediction
strategy, we first need a classifier that separates clear sky images from
cloudy ones.

We use a data set consisting of \(7000\) sky images labeled manually, from
which 75\% serve as a training set and the remaining sky images as a testing
set.  The classifier is a single hidden layer neural network that takes as
inputs the PCNP features extracted from the raw sky images and outputs two
probability scores corresponfding to the two labels. We use the sigmoid
function as neurons' activations. The neural network is trained using
Limited-memory BFGS algorithm which is known to converge to better solutions on
small data sets. Table~\ref{tab:classifier_summary} outlines information about
the classifier. 

\begin{table}[!t]
  \centering
  \caption{Neural network parameters summary for the sky image classifier}
  \label{tab:classifier_summary}
  \begin{tabular}{| l | l |}
    \hline
    \textbf{Parameter} & \textbf{Value} \\
    \hline
     Features & \(\text{PCNP}1, \text{PCNP}2, \dots, \text{PCNP}64\) \\
     Target & sky class: ``clear'' or ``cloudy'' \\
     Number of hidden layers & 1 \\
     Number of hidden units & 27 \\
     Activation function & sigmoid \(f(x) = \frac{1}{1 + e^{-x}}\) \\
     \(l_2\) regularization parameter & \(0.01\) \\
     Optimizer & Limited-memory BFGS \\
    \hline
  \end{tabular}
\end{table}

\section{GHI estimation}
In this section, we estimate the transfer function between the extracted PCNP
features and the measured GHI. The data set we use consists of 256 PCNP
features (i.e. \(k = 256\), with each feature representing the count of one of
the super-pixels, and a target variable which is the GHI. The data set counts
\(70000\) records after removing elements having a GHI of \(0\).

\begin{figure}[!t]
  \centering
  \includegraphics[width=0.5\textwidth]{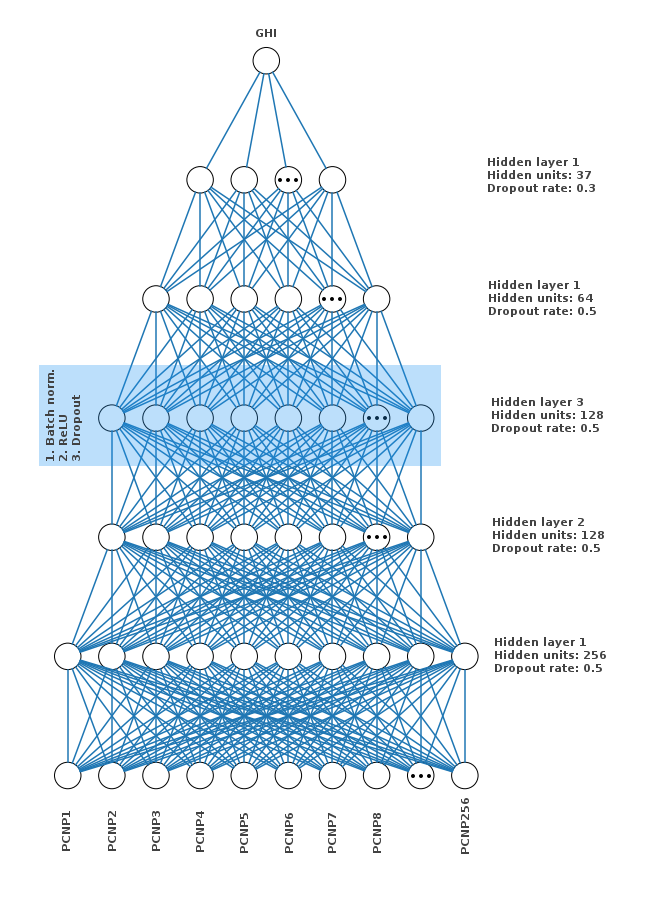}
  \caption{Architecture of the deep neural network}
  \label{fig:dnn_arch}
\end{figure}

Our modeling approach utilizes a deep neural network. The model's architecture
consists of five hidden layers \(L_1\), \(L_2\) \dots, \(L_5\) having \(256\),
\(128\), \(128\), \(64\) and \(37\) hidden units, respectively. We use as
activation function the Rectified Linear Unit \(\text{ReLU}(x) = \max(0, x)\)
which gives better results than the hyperbolic tangent \(\tanh(x) = \frac{e^x -
e^{-x}}{e^x + e^{-x}}\) usually used in regression problems.  Dropout
regularization is applied to each hidden layer of the network during the
training phase, which helps prevent over-fitting by reducing hidden units
co-adaptation. This technique gives major improvement to our model over
penalizing the weights with the \(l_2\) regularization. We also apply batch
normalization to each hidden layer in order to maintain the mean activation
close to \(0\) and the activation standard deviation close to \(1\). This has
the effect of accelerating the training phase by allowing the use of a higher
learning rate. It should be pointed out that the PCNP features were
preprocessed by scaling them to zero mean and unit standard deviation.
Figure~\ref{fig:dnn_arch} depicts our deep neural network's architecture.

The model is trained using Stochastic Gradient Descent algorithm with Nesterov
momentum, which is better suited for our large data set.  The hyper-parameters
of the deep neural network such as the number of hidden layers and neurons, the
activation function and the regularization parameters are determined using
10-fold cross validation. Table~\ref{tab:dnn_summary} summarizes the parameters
of our model.

\begin{table}[!t]
  \centering
  \caption{Deep neural network parameters summary}
  \begin{tabular}{| l | l |}
    \hline
    \textbf{Parameter} & \textbf{Value} \\
    \hline
     Features & \(\text{PCNP}1, \text{PCNP}2, \dots, \text{PCNP}256\) \\
     Target & GHI \\
     Number of hidden layers & 5 \\
     Number of hidden units & \(256\), \(128\), \(128\), \(64\) and \(37\) \\
     Activation function & \(\text{ReLU}(x) = max(0, x)\) \\
     Dropout rates & \(0.5\) and \(0.3\) for the last hidden layer \\
     Optimizer & Stochastic gradient descent \\
     Optimizer learning rate & \(0.01\) \\
     Optimizer momentum & Nesterov with parameter \(0.9\) \\
    \hline
  \end{tabular}
  \label{tab:dnn_summary}
\end{table}

\section{Results and discussion}
The estimation of the transfer function between the PCNP features and the GHI
shows on the testing set a coefficient of determination \(R^2\) of \(0.95\).
The corresponding mean absolute error on GHI estimation is about \(27\)
\(W/m^2\), where the measured GHI ranges from \(0\) to \(931\) \(W/m^2\). These
scores are validated using \(10\)-fold cross validation.

\begin{figure}[!t]
  \centering
  \subfloat[Clear sky conditions]{
    \includegraphics[width=3.3in]{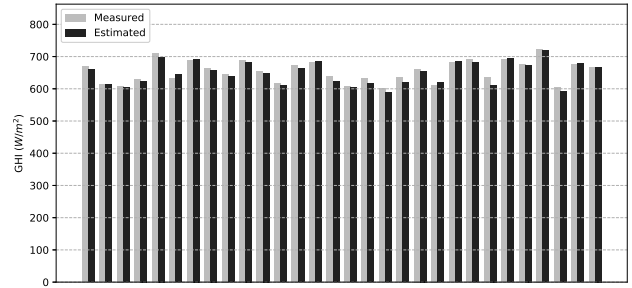}
    \label{fig:dnn_res_clear_sky}
  }
  \hfill
  \subfloat[Cloudy sky conditions]{
    \includegraphics[width=3.3in]{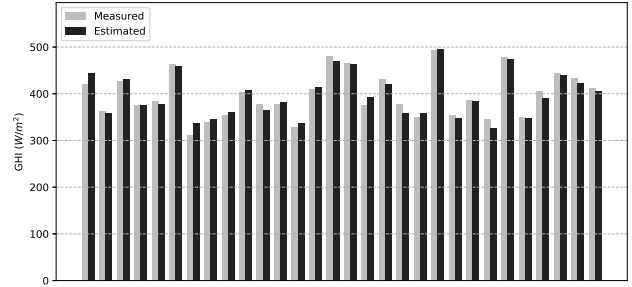}
    \label{fig:dnn_res_cloudy_sky}
  }
  \hfill
  \subfloat[Overcast sky conditions]{
    \includegraphics[width=3.3in]{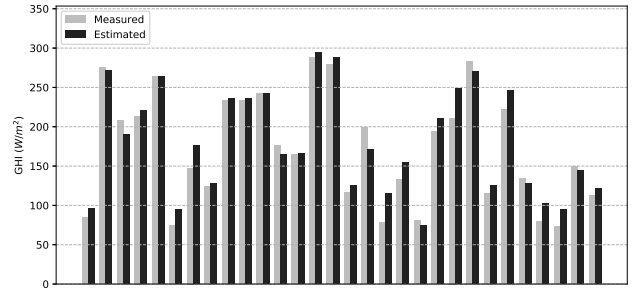}
    \label{fig:dnn_res_overcast_sky}
  }
  \caption{The estimated GHI (dark) versus the measured GHI (gray)
  in \(W/m^2\) for randomly selected entries from the testing set.}
  \label{fig:dnn_res}
\end{figure}

Figure~\ref{fig:dnn_res} shows the GHI output by the deep neural network versus
the real values acquired by the pyranometer for different weather conditions.
The entries are randomly selected from the testing set.

Concerning the neural network classifier, the metric we use to measure the
model's performance is the accuracy (i.e. the frequency with which predictions
matches labels).

\small
\[
  \text{Accuracy} = \frac{\text{True positives} + \text{True negatives}}
                         {\text{Number of predictions}}
\]
\normalsize

The classifier shows an accuracy of \(99.7\)\% on the testing set and is able
to distinguish between clear sky images having the glare surrounding the center
(c.f. Figure~\ref{fig:clear_sky}) and sky images having partial clouds. It is
worth mentioning that the model may be improved by making it able to detect
overcast sky conditions as well.

\vspace{5mm}

\section{Conclusion and future work}
In this paper, we presented a new method for modeling the relationship between
hemispheric sky images and their corresponding surface solar irradiances.  We
first used mini-batch \(k\)-means clustering in order to reduce the size of sky
images. The extracted PCNP features were used an inputs to train a deep
learning neural network to predict the irradiance associated with each sky
image. A real dataset was used to illustrate the merits of the proposed method.
As a future work, we will investigate how the extracted PCNP features can be
used to forecast intra-hour variability of the irradiance in the case of cloudy
and overcast sky conditions.

\section*{Acknowledgment}
This research work was supported by the Institute for Research in Solar Energy
and New Energies (IRESEN) and the United States Agency for International
Development (USAID). The authors would like to thank Yinghao Chu, Hugo T.
C. Pedro, Lukas Nonnenmacher, Rich H. Inman, Zhouyi Liao and Carlos F. M.
Coimbra for providing the data set used in this paper.

\vspace{5mm}

\bibliographystyle{utils/bibtex/IEEEtran}
\bibliography{utils/bibtex/IEEEabrv,src/references}

\end{document}